# Building a Stochastic Dynamic Model of Application Use

**Peter J. Gorniak**
Department of Computer Science
University of British Columbia
Vancouver, B.C., Canada, V6T 1Z4
pgorniak@cs.ubc.ca

**David Poole**
Department of Computer Science
University of British Columbia
Vancouver, B.C., Canada, V6T 1Z4
poole@cs.ubc.ca

## Abstract

Many intelligent user interfaces employ application and user models to determine the user's preferences, goals and likely future actions. Such models require application analysis, adaptation and expansion. Building and maintaining such models adds a substantial amount of time and labour to the application development cycle. We present a system that observes the interface of an unmodified application and records users' interactions with the application. From a history of such observations we build a coarse state space of observed interface states and actions between them. To refine the space, we hypothesize substates based upon the histories that led users to a given state. We evaluate the information gain of possible state splits, varying the length of the histories considered in such splits. In this way, we automatically produce a stochastic dynamic model of the application and of how it is used. To evaluate our approach, we present models derived from real-world application usage data.

## 1 Introduction

Artificial Intelligence supplies a vast set of tools to be applied to the design of intelligent user interfaces. While our previous research (Gorniak 1998) as well as countless other projects sample indulgently from this set and often produce quite impressive results in their own environments (for example, see (Horvitz, Breese, Heckerman, Hovel and Rommelse 1998) and (Albrecht, Zukerman, Nicholson and Bud 1997),) there emerge some new challenges when one attempts to apply these results to a new application.

1. The research results often do not transfer easily to a new application.

2. The actual implementation used in the research relies upon a modified application. This modification is usually non-trivial, time-consuming to repeat and increases application complexity.

3. Researchers work from various, often hand-crafted application models. In addition to the application building work in 2, an application designer needs to specify such a model. This process is often not straightforward and may rely on empirical data from user trials. An application designer's primary task does not include designing such a model, and thus the task seems an added difficulty to him or her. Also, the model needs to be updated and will tend to lag behind the application during maintenance.

We are currently addressing these problems by investigating how much knowledge can be extracted from a user's interaction with an application without any prior information about the application's purpose or structure and without any modifications to the application (somewhat in the spirit of (Lieberman 1998).) We hypothesize that enough knowledge can be extracted to yield a detailed model of the application and its user. This model can then serve both as a knowledge source for other algorithms as well as provide a context under which to unite methods. Specifically, we endeavour to construct a detailed state space together with a stochastic policy for the user's behaviour and in this way describe the application structure and the user traversing it. An intelligent user interface, for example a help system, can access information about the user's context and goals from our model, and thus tailor its presentation to the user's needs. Most importantly, we set out to construct this model without modifying the application and without running explicit user trials. In fact, the system we present works as a wrapper to the Java runtime environment and is able to model the use of any Java application run on the system without customization of either the application or our system.

Some other application independent user models build no application model at all, and thus do not provide any automatic analysis of the application. They perform worse in cases where such application knowledge boosts performance, such as future action prediction (Davison and Hirsh 1998). Others stop early on in their analysis and subsequently rely on application specific knowledge (Encarnacao 1997). We are not aware of other work that attempts to identify the current state of application and user without any knowledge or modification of the application. Related in the model building area is the next action prediction work for web precaching by Zukerman, Albrecht and Nicholson (1999). They work with combinations of simple Markov models and based on request counts and do not identify the system's state in more detail. Our goal is also akin to some work in the areas of data mining and specifically clustering. For example, Cadez, Heckerman, Meek, Smyth and White (2000) cluster users based on their web page request patterns for visualization purposes, but they do not build a detailed stochastic dynamic model like ours.

Let us view the user as an agent. Our assumption is that we can and have observed this agent acting in an environment, namely using an application. Artificial Intelligence concerns itself with agents acting in environments and worries about what decisions such agents should make. A common approach to such a problem consists of phrasing it in terms of states and actions between states and coming up with a policy that, perhaps stochastically, dictates which actions to take in which states (Boutilier, Dean and Hanks 1999). We are faced with the opposite problem: we see an agent acting in an environment and want to model the agent's decision process. We assume that the agent acts according to a policy. Each action is the result of some (possibly stochastic) function of what the agent observes and the agent's belief state. Our goal is to determine this policy and the state space to which it applies.

Previously, we have shown this approach to perform exceedingly well in predicting future user actions (Gorniak and Poole 2000). In that research we identified the user's state implicitly by finding the longest sequences in observed history that match the actions the user just performed, similar to (McCallum 1996). We ranked possible future actions according to the lengths of these matches. Our goal in the research presented here is to explicitly identify the states of user and application. This results in a detailed model of how one or many people are using the application. Such a model can help application designers analyse their applications and augment them with intelligent extensions. For example, it is easy to learn about user behaviour and find unexpected consequences of a design decision. Or, in building an intelligent help system, the designer can query our system for the user's current state and context, as well as our prediction for the user's future actions and then tailor the help to that scenario. Note that while we lean on some techniques from classification algorithms, we are not dealing with a standard machine learning problem here. Our emphasis lies on a humanly readable and usable model. Therefore, we cannot offer a numerical performance measure of how well our system performs its task. Instead, we illustrate that the model we infer for real application usage data provides a useful foundation for application analysis and extension.

In Section 2 we describe the motivation behind our state identification algorithm and compare it to its implicit sibling, ONISI (On-Line Implicit State Identification (Gorniak and Poole 2000)). Section 3 describes the Java implementation of the work presented here. This implementation works as a wrapper to existing Java applications and is able to record their interfaces states as well as user actions without modifications to the original application. Section 4 discusses the performance of this algorithm on real user data from an example application and analyses the resulting application model. Finally, Section 5 concludes and points to future work.

## 2 Explicit State Identification Algorithm

Our implementation (see Section 3) supplies us with a record of observations of the application's interface states and the user's actions that lead from one observed interface state to another. That is, our recorded history consists of sequences of state observation/user action pairs. The observed states are often very coarse in that they are nowhere close to fulfilling the Markov assumption. Users tend to take very different courses of actions from them according to different goals. Overall, our approach to automatically deriving a refined state space (as with ONISI in (Gorniak and Poole 2000)) consists of identifying behavioural patterns that the user engages in and that predict future user actions well. However, to predict the next user action ONISI was given the history sequence that just occurred and it proceeded to find long matches to it in recorded history, under the (correct) assumption that users normally continue such patterns in identical ways. OFESI (Off-line Explicit State Identification), the algorithm discussed here, has a different goal. Rather than striving to predict the next action accurately, its purpose is to identify meaningful states that explain overall user behaviour. As a consequence, ONISI considers a single long match of a behavioural pattern an important predictor, but OFESI needs to group identified patterns into sets that delimit distinct user states.

This state refinement problem can be viewed as a classification task: given the occurrences of a state in recorded history, the action sequences that precede them, and the actions that follow them, how should we group the sequences such that the groups give us as much information as possible about what action will occur after the state? This problem sounds much akin to the problem of picking an attribute to split on in building a decision tree using ID3 (Quinlan 1986). However, the natural attributes to use in splitting a state are the action sequences preceding it. These attributes are many-valued, producing a split into a large number of substates. Instead, we would like to split the state into as many states 'as make sense', that is, as are useful in capturing possible user intentions when reaching the observed interface state. We need to dynamically construct attributes with fewer values to predict substates.

We choose for OFESI to perform hierarchical binary splits of a state according to how much information such a split yields about the actions taken from it, and according to how many instances of such actions the new substates explain. Grouping the preceding action sequences in such a way supplies us with a new attribute with values that identify relevant substates well.

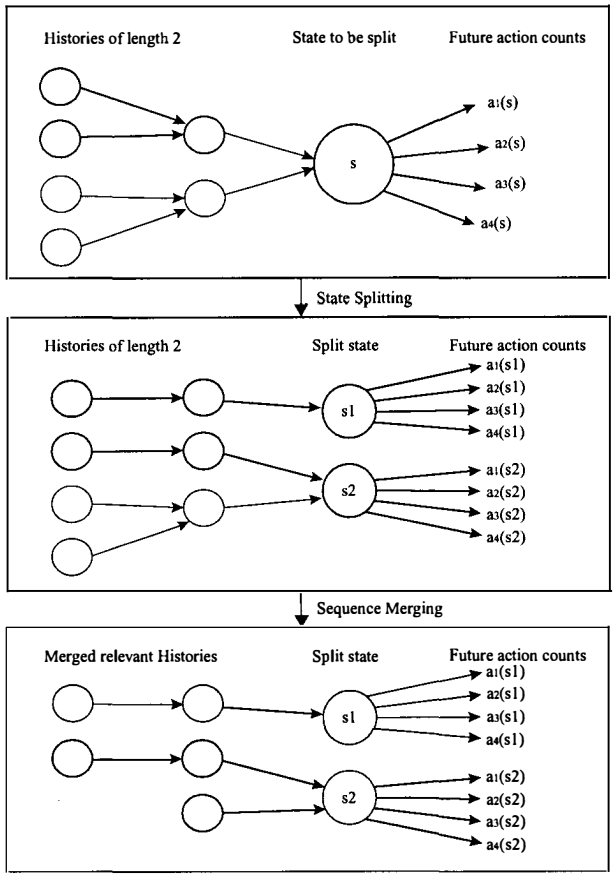

Figure 1: State Splitting

Figure 1 depicts the schema OFESI employs in splitting a state. We initially consider all distinct fixed length history sequences that have preceded an interface state in recorded history. In the following discussion, a state is a set of such sequences together with the observed interface state they lead to. The length of these sequences is an input to the algorithm. We discuss the choice of length in Section 4. History also supplies us with a distribution over next actions that users chose in the state. In fact, each preceding action sequence carries such a distribution, their sum being the state's action distribution. We now split the set of preceding history sequences into two. For the resulting sets, we sum the next action distributions of their constituent members to arrive at two distributions for the newly created substates. In turn, we may choose to split the new substates again in a hierarchical fashion. There is likely to be some redundancy amongst the resulting sets. For example, if all preceding sequences that end in the same action are grouped into one of the substates, that single action is enough to identify all these sequences. Thus, OFESI's last step examines the sequence sets for such redundancies and replaces groups of sequences with a shorter sequence wherever possible.

The main question now is: how can we split the set of preceding fixed length sequences into two, such that the resulting sets convey as much information as possible about the next action? Just as in building a decision tree, we evaluate possible sets according to their information gain. For a set of history sequences defining a state $s$ the information needed to fully predict the next action if there are $n$ actions possible overall is

$$I(s) = -\sum_{i=1}^{n} P_s(a_i) \log P_s(a_i)$$

where $P_s(a_i)$ is the probability with which action $i$ occurs in state $s$, i.e. the number of times the action occurs in this state in history divided by the number of times the state occurs in history (Shannon and Weaver 1949).

A new subset $s_1$ of $s$ leaves us with a remaining information need of

$$R(s_1) = -\sum_{j=1}^{n} P_{s_1}(a_j) \log P_{s_1}(a_j),$$

for that part of the original state, so the split of $s$ into $s1$ and $s2$ yields an information gain of

$$G(s, s_1, s_2) = I(s) - P(s_1)R(s_1) - P(s_2)R(s_2),$$

where $P(s_1)$ is the probability with which the predictions grouped into substate $s_1$ occur. We need to split the set of action sequences into two subsets such that this value is maximized.

We use a form of stochastic local search to optimize information gain of the subset split (Hoos 1998). Specifically, we initialize the search by splitting the history sequences leading to the interface state into two random subsets. Each search step moves exactly one sequence from one set to the other. To perform a step we pick the move that increases information gain the most with probability $p$ and pick a random move with probability $1 - p$. We only move attributes that have not been moved for at least $k$ steps, and reset the search process if information gain has not improved for $m$ search steps. We discuss the parameter settings for this search in Section 4.4.

```
//States are represented by sets of
//action sequences leading to single
//interface states.
Given settings for Amin, Gmin,
     sequenceLength;
OFESI() {
   For each interface state s {
      Let resultingStates = {};
      Find all actionSequences of
         length sequenceLength
         that precede s;
      Call OFESI-split-state(
         actionSequences,
         resultingStates);
      Replace s by substates found
         in resultingStates;
   }
}
OFESI-split-state(actionSequences,
               resultingStates) {
   Let {gain, substate1, substate2} =
      OFESI-binary-split(
         actionSequences);
   If gain > Gmin
      AND the number of actions
         predicted by each of
         state1 and state2
         is greater than Amin {
      Call OFESI-split-state(state1,
         resultingStates);
      Call OFESI-split-state(state2,
         resultingStates);
   } else {
      Append actionSequences to
         resultingStates;
   }
}
```

Figure 2: The OFESI Algorithm

```
Given settings for searchSteps,
     p, k, m;
OFESI-binary-split(actionSequences) {
   Randomly split actionSequences into
         state1, state2;
   Let gain = G(actionSequences, state1,
               state2);
   Let bestGain = gain;
   Let stagnantCount = 0;
   Let bestSplit = {state1, state2}
   For searchSteps number of times {
      With probability p {
         Move action a that maximises
               gain improvement
               when moved between
               sets and hasn't been
               moved for k steps;
      } else {
         Move random action a;
      }
      Let gain = G(actionSequences,
               state1, state2);
      If gain > bestGain {
         Let bestGain = gain;
         Let bestSplit = {state1,
            state2};
         Let stagnantCount = 0;
      } else {
         stagnantCount++;
      }
      if(stagnantCount > m)
         Randomly split
            actionSequences into
            state1, state2;
   }
   return {bestGain, bestSplit};
}
```

Figure 3: Binary State Split Algorithm

OFESI accepts the best split if it yields an information gain of at least $G_{min}$ and if for each set the sum of the action instances that set predicts is at least $A_{min}$. This restricts splits to those that still yield a reasonable amount of information about the actions taken from a state and avoids splits that do yield information, but create substates that explain an insignificant amount of actions. Both parameters cannot be optimized in any objective sense, but rather depend on one's goals in using OFESI. We discuss their impact in Section 4.3.

Upon a successful split, OFESI now considers each substate as the new state to be split and continues splitting in this hierarchical fashion as long as each split fulfills the $G_{min}$ and $A_{min}$ restrictions. Figures 2 and 3 outline the complete OFESI algorithm without the final sequence merging step.

## 3 Implementation

Java's reflective capabilities and dynamic loading strategy make the language a prime candidate for an application independent approach (JDK 1998). It allows not only inspection of a structure of known visual components, but it can also inspect unknown components for state information. Java and JavaBeans introduced standard naming conventions for object methods. For example, *isVisible()* returns the visibility status of visual components, whereas *getEnabled()* returns whether they are currently useable. Components derived from standard Abstract Window Toolkit components inherit these methods automatically, and other components should define them. Java's Reflection mechanism, on the other hand, allows one to check whether a given object includes one of these state-revealing methods, and lets one call this method without knowing the object's class. Finally, Java's dynamic loading of classes rids the developer of needing to link with or even know about classes that will be present at runtime. Using these tools, one can establish the user interface state of an application built using Java at runtime by dynamically linking into its code, examining the methods available in its objects and calling the methods relevant to the interface state. This process requires no modification of the targeted application at all.

The system used for the experiments presented below runs as a wrapper to a Java application. Before it starts the application, it hooks itself into the application's event queue and thus sees all event activity within the Java Abstract Window Toolkit and components derived from it. It intercepts each such event that it considers an action (such as a button being pressed or a window closed) and records the observed state of the application's interface before and after the event occurs. In this way, this system establishes a state space of interface observations as a person uses the application and records a history consisting of actions and visited states at the same time.

The applications [1] under consideration here are educational AI applications. They were written to help undergraduate university students learn concepts in Artificial Intelligence. One application familiarizes the student with search problems and algorithms, the second deals with constraint satisfaction problems and the third demonstrates backpropagation neural network learning. In each, the student has the option to either load an example problem or to create his or her own problem by drawing a graph. He or she can then switch to a problem solution mode and step through the various algorithms at different levels of detail. The students used these applications to solve homework problems for an introductory AI course they were taking. Most of the assignment questions referred to a supplied example problem, so the students tended to explore the problem creation facilities of the applications less than their solving functionality. The following discussion and results focus mainly on the application for search algorithms.

## 4 Results and Discussion

There exists no obvious user-independent performance measure for the system presented here. Its usefulness depends on the goals of the person employing the system, be it to debug an existing application, to design additional application components or to simply perform a study of application usage. We have evaluated the implicit version of our state identification approach to predict future user actions in (Gorniak and Poole 2000) and we are currently working to apply the explicit version presented here to another problem that can benefit from explicit state identification, namely that of deriving structure for Hidden Markov Models (Rabiner 1989). In the following sections, we present the model OFESI derives for the search application and argue that it captures significant features of user behaviour.

---

[1] The applications can be found at *http://www.cs.ubc.ca/labs/lci/CIspace/*.

### 4.1 Observed State Space

Figure 4 shows the state space of interface states our system observes from users of the search algorithm application. The figure represents the space exactly as recorded, except for that we have given the states meaningful names. Reflecting the division of the application into two modes, problem creation and problem solution mode, the graph exhibits two distinct components. The right hand component corresponds to problem solution mode, whereas the left hand one corresponds to problem creation mode. The students were mainly using the application to solve problems that were given to them, so we recorded significantly more data for problem solution mode. The following discussion therefore focusses on the right hand subcomponent of Figure 4.

In this component, we see two distinct ways of examining search algorithms using this application. Students can either step through a problem using a search algorithm, or they can show the result of the algorithm given the problem. At most times they can reset the search, which transports them back to the **Problem Solution** start state. During stepping, they can still ask to be shown the result at any time. **Show Result** and **Goal Node Reached** are the states in which dialog boxes are shown. We can distinguish two versions of these states, one in which the student has stepped previously, and one in which the student asked to see the result directly from the **Problem Solution** state. While this graph is interesting, it tells us little about whether a student will choose to step or examine the result in **Problem Solution** state. We now demonstrate how OFESI splits states to give us exactly that information.

### 4.2 State Splitting Results

First, let us examine the state **Stepping** and how OFESI splits it. Table 1 lists the original next action distribution of this state.

We see that there are three main actions users choose from this state: They either step, fine step, or reset the search. Intuitively, we would like to split the state into three substates, each predicted by an appropriate set of history sequences leading to it. Tables 2 and 3 show a substate OFESI suggests when considering history sequences of length one by giving the action sequences predicting the state and the action distribution in the state, respectively.

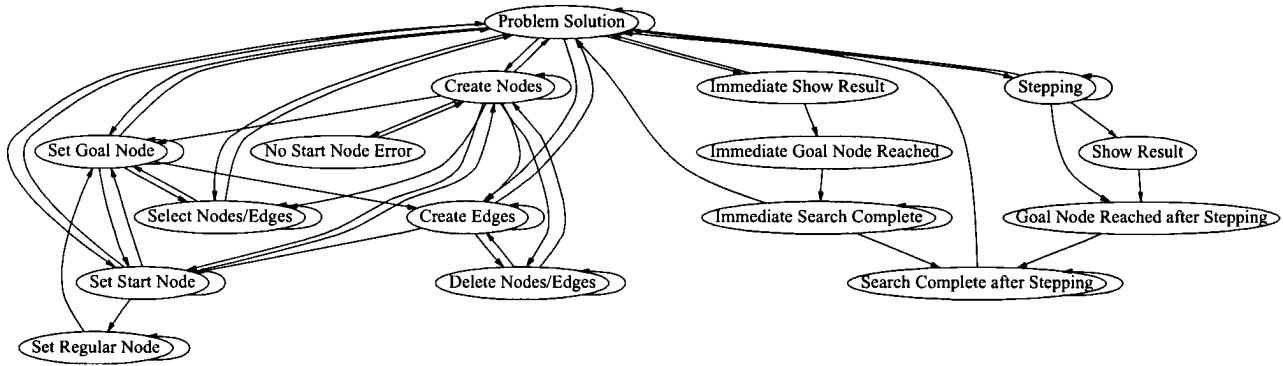

Figure 4: Original Interface States Observed in Search Application

Table 1: Original Action Distribution for **Stepping**

| Action | Count |
| --- | --- |
| Show Frontier | 5 |
| Reset Search | 54 |
| Step | 738 |
| Display Node Heuristics | 1 |
| Show Result | 6 |
| Display Edge Costs | 1 |
| Fine Step | 393 |

This substate is a result of the initial binary split of **Stepping** OFESI performs, which splits the state into one substate predominantly predicting the Step action and another predicting the Fine Step action.

Table 2: Sequences Defining **Stepping** Substate

| Previous State | Previous Action |
| --- | --- |
| Problem Solution | Step |
| Stepping | Display Edge Costs |
| Stepping | Step |

This substate makes sense, judging from the intuition behind the action sequences that predict it (a user that stepped before is likely to step again), and from the next action distribution that is dominated by the Step action. This substate will not be split again, but OFESI does split its dual substate in a hierarchical call into substates that predict the Fine Step action and the Reset Search action (it is predicted by the last state having been **Goal Node Reached after Stepping**.) As run here, with $G_{min} = 0.15$, $A_{min} = 10$ and an action sequence length of one, OFESI suggests exactly three states. We do not include the details for the other two substates for space reasons.

The choice of action sequence length constitutes a trade off between computational and explanatory complexity on the one hand and explanatory power on the other hand. That is, short action sequences (say, of length one) are easy to read and there are few of them, whereas there tend to be exponentially more sequences with each additional action considered. More longer sequences allow us to split the state at least as well, and usually better, than few shorter sequences, but due to their number the splitting process is computational more expensive and the resulting substates are hard to interpret based on the action sequences (they are still often easily interpretable from the actions they predict.) At the same time, overfitting may occur with longer action sequences in the sense that patterns peculiar to the training history may be used to identify substates.

Table 4 shows the action distribution for the substate predicting the Step action as derived by OFESI run with sequence length four. It is clear that the split is cleaner - there are no more Fine Step actions predicted by this state, and more Step actions predicted. We refrain from including the unwieldy set of action sequences that predicts this state.

Table 3: Next Action Counts for **Stepping** Substate (Table 2)

| Action | Count |
| --- | --- |
| Show Frontier | 4 |
| Reset Search | 16 |
| Step | 733 |
| Display Node Heuristics | 1 |
| Show Result | 5 |
| Fine Step | 2 |

Table 4: Refined Next Actions Counts

| Action | Count |
| --- | --- |
| Show Frontier | 2 |
| Reset Search | 13 |
| Step | 736 |
| Display Node Heuristics | 1 |
| Show Result | 4 |

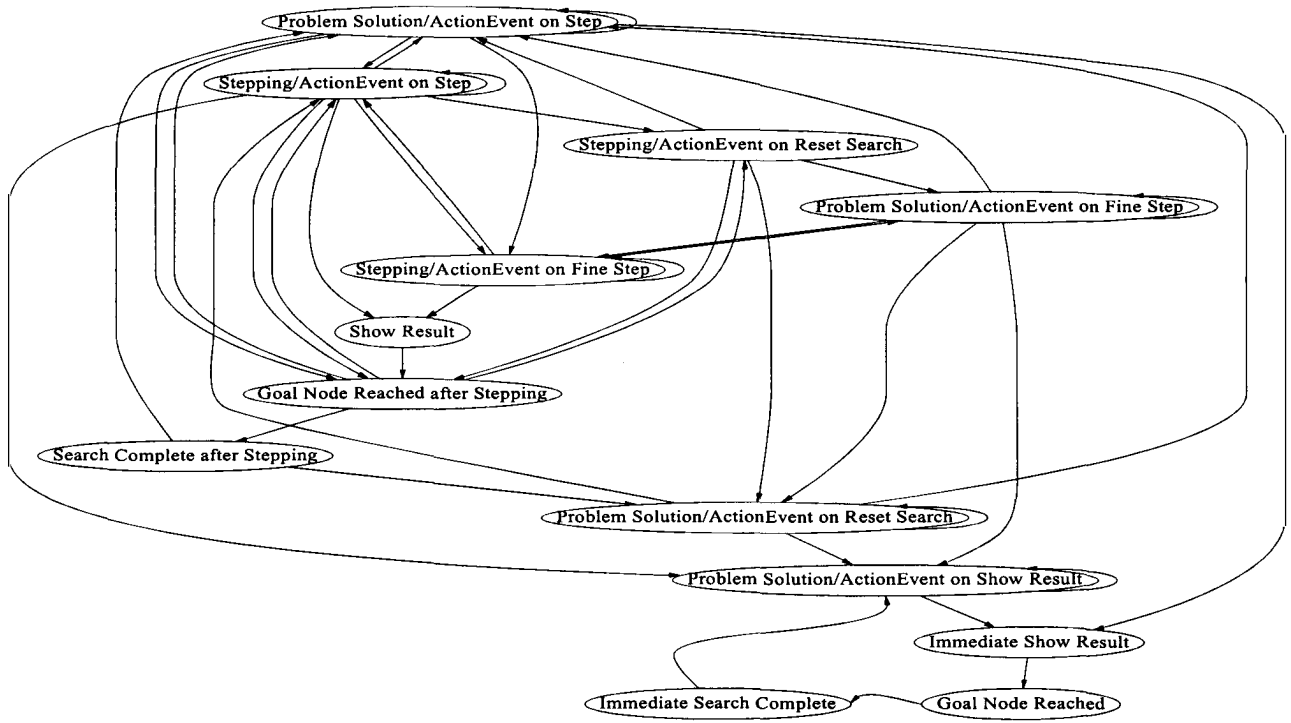

Figure 5: Split State Subspace for Search Application

Upon examining this set we find that in essence OFESI was able to take into account features like that if the last action was Step, but the three before that were Fine Step, the user is likely to choose Fine Step next.

### 4.3 Refined State Space

Figure 5 shows the state space after running OFESI on the right hand component of state space shown in Figure 4. States are labelled by their original name followed by the most frequently occurring action in their next action distribution if they are substates of the originally observed interface states. OFESI was run with $G_{min} = 0.15$, $A_{min} = 10$ and an action sequence length of 5. These parameters can be set differently according to one's goal in running OFESI. Generally, lower settings of $G_{min}$ will produce more states, but states at lower levels of the hierarchical split will tend not to contribute much to distinguishing between possible next actions. Lower settings of $A_{min}$ will also produce more states, but these states will tend to distinguish actions that occur less often. First, notice the three substates of the **Stepping** state as discussed in Section 4.2. In the same fashion, OFESI splits **Problem Solution** into four substates, according to whether the user is likely to Step or Fine Step through the problem, to ask to be shown the result or to reset the search. The last of these **Problem Solution/ActionEvent on Reset Search** is an artifact of the user interface design. The problem solution process is in its initial state if the user is in **Problem Solution** state, but users ask for the process to be reset both after just having reset it and after switching search strategies or search options (as obvious from the action sequences OFESI attaches to the new substate.) Clearly, it is due to a flaw in the interface that users engage in this behaviour. In general, the state space shown in Figure 5 presents much more detailed model of application use, and its states yield far more predictive power than the originally observed coarse interface states.

### 4.4 Information Gain Optimization

As mentioned before, we employ stochastic local search to optimize information gain when splitting a state into two substates. The complexity of this search largely depends on the length of the action sequences considered. With sequences of length one, there are only few items to be grouped and most parameter settings find the optimal solution quickly. With longer sequence the number of items in the groups tends to grow exponentially, and the search quickly becomes more complex. However, we found that with sequences of length five the search rarely find better solutions after more than 500 steps. In addition, the improvements in later steps tend to be much smaller than in earlier ones and we do not necessarily care to find the optimal split as long as we find a very good one, so running the search to 500 steps, with a probability of random steps of 0.1 and restarting after the solution has not improved after 80 steps proved sufficient for our purposes. We do not claim that these are in any way the optimal parameters, but they appear to work well enough in practise.

## 5 Conclusion and Future Work

We have presented a system that observes user actions and application interface states from an unmodified application. It deduces a coarse state space that the user is traversing from this observed history and a stochastic policy that describes the user's behaviour. Together, these form a stochastic dynamic model of application use. Each interface observed state has an associated next action distribution that potentially includes high frequencies for several actions. To enhance the model, we use OFESI to split states in a hierarchical fashion by optimizing the information gain on the original next action distribution that such a split yields. We have shown this algorithm to split states recorded in user trials into substates of good predictive power. These substates form a new, more informative state space of the use of the application, and supply us with a more decisive policy.

While the state spaces derived by OFESI are those one would wish for, we intend to prove the usefulness of our automatic application and user modeling strategy by augmenting one of our applications with a component that uses the model's information. We also intend to build an application design and analysis tool that uses OFESI to produce a state space of an existing application and helps the application designer in evaluating past design decisions based upon real application usage. In addition, an application designer could add information to the state space, for example by giving states names or by grouping states into contexts, and could then interface to an API we provide to access information about the current state and context as well as future contexts and goals of the user.

Finally, there is another area of research that could take advantage of a general way to infer meaningful states from a sequence of observations: Hidden Markov Models (Rabiner 1989). We are currently investigating whether OFESI as presented here can help in finding the optimal number of states for a Hidden Markov Model and give hints as to which observations each state should account for and where in the model it belongs.

## References

Albrecht, D. W., Zukerman, I., Nicholson, A. E. and Bud, A.: 1997, Towards a bayesian model for keyhole plan recognition in large domains, *User Modeling: Proceedings of the Sixth International Conference, UM97*.

Boutilier, C., Dean, T. and Hanks, S.: 1999, Decision-theoretic planning: Structural assumptions and computational leverage, *Journal of AI Research* 11, 1–94.

Cadez, I., Heckerman, D., Meek, C., Smyth, P. and White, S.: 2000, Visualization of navigation patterns on a web site using model based clustering, *Technical Report MSR-TR-00-18*, Unversity of California, Irvine.

Davison, B. D. and Hirsh, H.: 1998, Predicting sequences of user actions, *Technical report*, Rutgers, The State University of New York.

Encarnacao, L.: 1997, *Concept and Realization of intelligent user support in interactive graphics applications*, PhD thesis, Eberhard-Karls-Universität Tübingen, Fakultät für Informatik.

Gorniak, P. J.: 1998, Sorting email messages by topic. Project Report.

Gorniak, P. J. and Poole, D.: 2000, Predicting future user actions by observing unmodified applications, *Proceedings of the 17th National Conference on Artificial Intelligence, AAAI-2000*.

Hoos, H. H.: 1998, *Stochastic Local Search – Method, Models and Applications*, PhD thesis, Technische Universität Darmstadt.

Horvitz, E., Breese, J., Heckerman, D., Hovel, D. and Rommelse, K.: 1998, The lumiere project: Bayesian user modeling for inferring the goals and needs of software users, *Uncertainty in Artifical Intelligence, Proceedings of the Fourteenth Conference*.

JDK: 1998, *Java Development Kit Documentation*. URL: *http://java.sun.com/products/jdk/1.1/docs/*

Lieberman, H.: 1998, Integrating user interface agents with conventional applications, *Proceedings of the International Conference on Intelligent User Interfaces, San Francisco*.

McCallum, A. R.: 1996, Instance-based state identification for reinforcement learning, *Technical report*, University of Rochester.

Quinlan, J.: 1986, Induction of decision trees, *Machine Learning* 1, 81–106.

Rabiner, L.: 1989, A tutorial on hidden markov models and selected applications in speech recognition, *Proceedings of the IEEE*, Vol. 77(2).

Shannon, C. and Weaver, W.: 1949, *The Mathematical Theory of Communication*, University of Illionois Press, Urbana.

Zukerman, I., Albrecht, D. and Nicholson, A.: 1999, Predicting users' requests on the www, *User Modeling: Proceedings of the 7th International Conference, UM99*.